\documentclass[10pt,twocolumn,letterpaper]{article}

\usepackage{iccv}
\usepackage{times}
\usepackage{epsfig}
\usepackage{graphicx}
\usepackage{amsmath}
\usepackage{amssymb}

\usepackage{authblk}
\makeatletter
\renewcommand\AB@affilsepx{, \protect\Affilfont}
\makeatother

\usepackage[breaklinks=true,bookmarks=false]{hyperref}

\iccvfinalcopy 

\def\iccvPaperID{****} 

\begin{document}

\title{Dance Dance Generation: Motion Transfer for Internet Videos}
\author[1]{Yipin Zhou}
\author[2]{Zhaowen Wang}
\author[3]{Chen Fang}
\author[2]{Trung Bui}
\author[1]{Tamara L. Berg}
\affil[1]{University of North Carolina at Chapel Hill}
\affil[2]{Adobe Research}
\affil[3]{ByteDance AI Lab}

\maketitle

\begin{abstract}
This work presents computational methods for transferring body movements from one person to another with videos collected in the wild. Specifically, we train a personalized model on a single video from the Internet which can generate videos of this target person driven by the motions of other people. Our model is built on two generative networks: a human (foreground) synthesis net which generates photo-realistic imagery of the target person in a novel pose, and a fusion net which combines the generated foreground with the scene (background), adding shadows or reflections as needed to enhance realism.
We validate the the efficacy of our proposed models over baselines with qualitative and quantitative evaluations as well as a subjective test.
\end{abstract}

\vspace{-.4cm}
\section{Introduction}
\label{introduction}
\vspace{-.1cm}

Imitation is a common everyday experience, from babies copying their parents' movements, to instructional videos and video games, people often learn new skills by mimicking others. For example, in dance classes or other sports, we mimic the movements of instructors, but it can take years to advance from amateur to professional level. What if you could skip those thousands of hours of practice and simply generate a video of yourself excelling in an extreme sport or instantly copying your favorite celebrities' dance steps?



\begin{figure}[t]
\begin{center}
\includegraphics[width=1.0\linewidth]{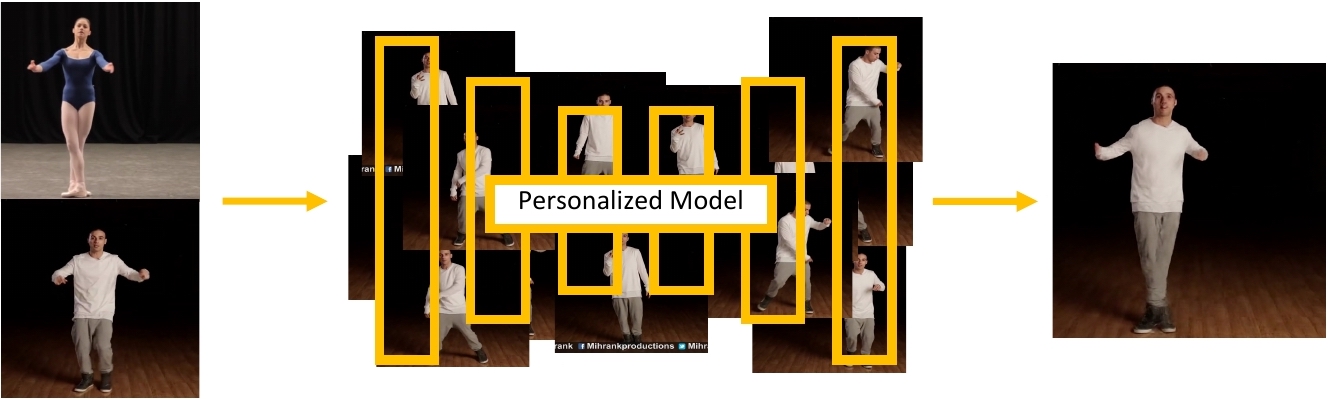}
\end{center}
\vspace{-.3cm}
   \caption{We train a personalized model for a target individual in an Internet video. This model can synthesize the target person in novel poses (right) from the input frame (bottom left) driven by a different individual (top left).}
\label{fig:overview}
\vspace{-.4cm}
\end{figure}

\begin{figure*}[t]
\begin{center}
\includegraphics[width=1.0\linewidth]{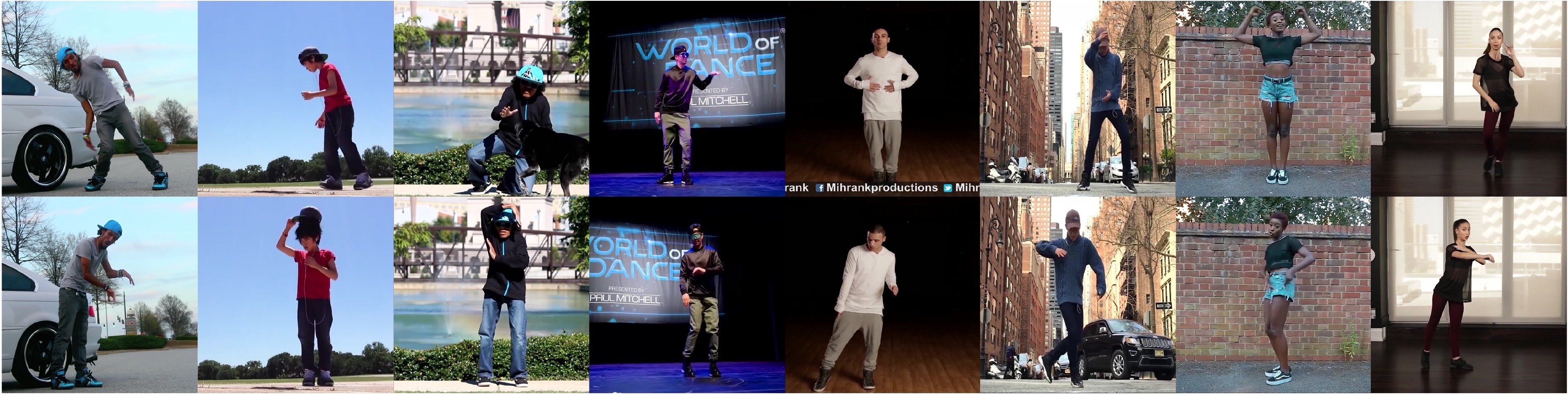}
\end{center}
\vspace{-.3cm}
   \caption{Example frames of personal videos from YouTube (from left to right are: Video\_01 to Video\_08).}
\label{fig:example_frames}
\vspace{-.4cm}
\end{figure*}

In this paper, we develop computational methods for imitating human movements. In particular, we train a model using one several-minute-long video of a target person and can then transfer any desired movement from a new reference video to the target person (maintaining appearance) as Fig.~\ref{fig:overview} shows. Previous work~\cite{pose_guide_2017,deformable_2018_CVPR,unseen_2018_CVPR,everybody_dance} investigates human pose transfer either under generalized settings (models trained across various individuals/scenes) or using single person lab-recorded videos. In this work, we explore personalized motion transfer on videos obtained from the Internet, such as YouTube. With such data pre-recorded in uncontrolled ways, our models are required to generalize well to novel poses, as we cannot ask a person on YouTube to demonstrate the entire range of poses we might want to generate. We show example frames of the Internet videos used as training data in Fig.~\ref{fig:example_frames}.

We formulate the motion transfer problem as a pose guided image translation task. Specifically, given an input frame (from the target person) and a desired pose, our model synthesizes the output as the target person in that new pose. Because our goal is to develop a general method that can handle unconstrained target videos from the Internet, which may contain moving background objects or unstable camera motions, we break our task down into two pieces that deal with the foreground (person) and background separately. 
In particular, our model is composed of two stages: the human synthesis stage which extracts body segments of the target person and then generates a photo-realistic image of the target in a new pose; and a fusion stage which combines the generated person with the extracted background scene. The latter fusion stage helps to remove small artifacts introduced from the human generation stage as well as adding shadows/reflections due to the interaction between the person and the scene. In addition, because we separate foreground and background processing, a side benefit of our method is the ability to place the generated person on a {\em new background} (something that cannot be achieved with whole frame generation methods).
To obtain high-quality synthesis results, we also apply temporal smoothing to generate temporally coherent movements and use hand, foot, and face landmarks to enhance local body details. We describe our method in Sec.~\ref{approaches}.
To evaluate the effectiveness of the proposed method, we conduct both numerical evaluations and human experiments on generated results in Sec.~\ref{experiments}.

In summary, our contributions include the following. 1) We demonstrate personalized motion transfer on videos from the Internet. 2) We propose a novel two-stage framework to synthesize people performing new movements and fuse them seamlessly with background scenes. 3) We perform qualitative and quantitative evaluations validating the superiority of our method over existing state-of-the-art.

\vspace{-.1cm}
\section{Related work}
\label{relatated_work}
\vspace{-.1cm}

\noindent {\bf Pixel to pixel translation:} Recent developments in generative models have started to make high quality photo-realistic image synthesis possible. Pixel to pixel translation frameworks are conditioned on input images or video frames and learn a mapping from the input domain to the same or different output domain. Image-to-image translation~\cite{pix2pix} was one of the first papers to present a general framework for handling a variety of translation tasks, such as edge to pixel or grayscale to color translation, using a conditional Generative Adversarial Network (GAN)~\cite{GAN} to learn mappings from paired data. 
CycleGAN~\cite{CycleGAN} further introduced a novel cycle consistency loss to learn domain translations without requiring paired image data. In follow-on work, Recycle-GAN~\cite{Recycle-GAN} added spatial and temporal constraints to enable video to video translation. Cascaded refinement frameworks~\cite{Qifeng_2017_ICCV,Qi_2018_CVPR} were designed to synthesize images from semantic layouts, achieving visually appealing generation. 
To further improve results, pix2pixHD \cite{pix2pixHD} proposed a multi-scale conditional GAN to synthesize high-resolution images from semantic labels, while spatio-temporal adversarial constraints were added to generate temporally consistent results for video to video synthesis~\cite{vid2vid}.

\vspace{+.1cm}
\noindent {\bf Motion transfer:} Generative human motion transfer learns a mapping from input images of a person to generate images showing the person in new poses. Various models have been proposed which are either conditioned on a target pose or map directly from poses/surface maps (without requiring an input human image) to the desired output image. Ma et al.~\cite{pose_guide_2017} use a two-stack U-Net~\cite{pix2pix} framework to synthesize images of people given an input image of the person and an arbitrary target pose. Siarohin et al.~\cite{deformable_2018_CVPR} solve the same pose transfer problem by applying novel deformable skip connections in the generative architecture.  Similarly, Balakrishnan et al.~\cite{unseen_2018_CVPR} generate a depiction of a person in a novel pose given an image of the person and the target pose under a framework 
with sub-modules to composite the transformed foreground layers with a hole-filled background. On the other hand, Chan et al.~\cite{everybody_dance} learn the mapping directly from detected poses to generate a target person's video by applying pix2pixHD~\cite{pix2pixHD} with temporal smoothing and face GAN structures. This works quite well in constrained video domain where models are trained on lab-made videos of a single person showing a large variety of poses. Neverova et al.~\cite{dense_pose_2018_ECCV} generate images based on an input person image and a surface map (dense 3D pose). They present a framework that combines the pixel-level prediction and UV texture mapping. Very recent work~\cite{Liu2018Neural} presents a method to transfer poses from a source video to the target person. Specifically, they reconstruct a 3D model of the person and train a generative model to produce photo-realistic frames based on images rendered with this 3D model.

\begin{figure*}[t]
\begin{center}
\includegraphics[width=1.0\linewidth]{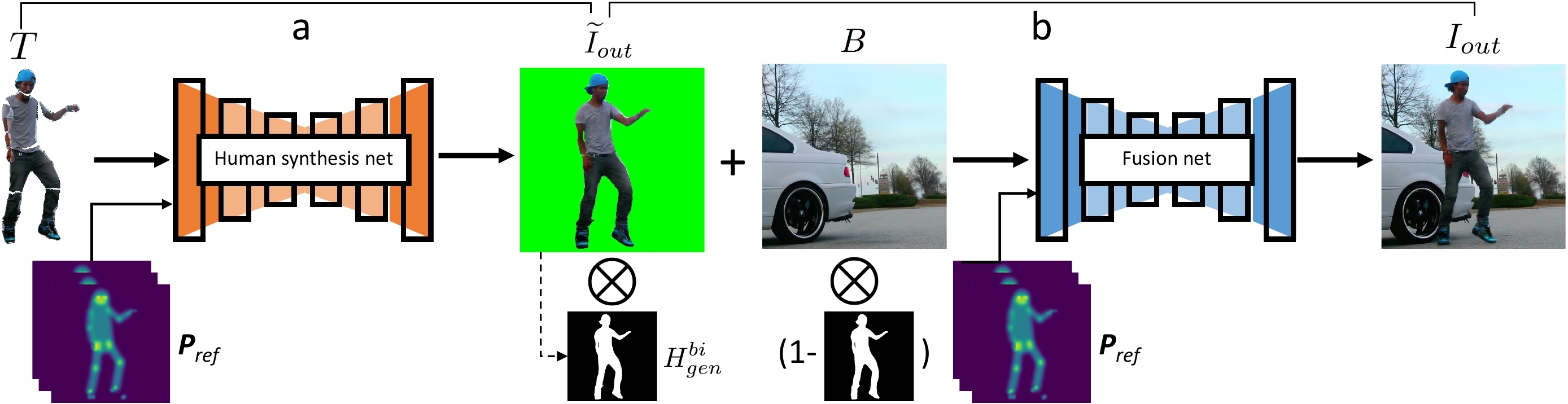}
\end{center}
\vspace{-.2cm}
   \caption{Two-stage motion transfer framework. (a) The human synthesis network takes transformed body parts $T$ as input to synthesize human image $\widetilde{I}_{out}$ on a green background, simplifying foreground $H_{gen}^{bi}$ extraction. $\textbf{P}_{ref}$ represents the aggregated pose maps. (b) The fusion network takes the combined foreground image $\widetilde{I}_{out}$ and fixed background $B$ as input and synthesizes the final output frame $I_{out}$.}
\label{fig:two-stage_framework}
\vspace{-.4cm}
\end{figure*}
\vspace{-.1cm}
\section{Approach}
\label{approaches}
\vspace{-.1cm}
We formulate the personalized motion transfer task as follows. Given an input frame $I_{in}$ and a reference frame $I_{ref}$ of size $H{\times}W$, as well as their associated human poses $P_{in}$ and $P_{ref}$, we would like to learn a mapping $\mathcal{F}$ that generates an output frame $I_{out}$ with the reference pose transferred to the target person in the input frame:
\begin{equation}
I_{out} = \mathcal{F}(I_{in}, P_{in}, P_{ref})
\end{equation}
where $I_{out}$ retains the same human appearance and background scene as $I_{in}$ while rendering with the reference pose, $P_{ref}$. During training, we sample random frame pairs from a personal video as our training data $(I_{in}, I_{ref})$.

We treat the human body part segments from $I_{in}$ as tangram pieces which are placed on a composition image according to the layout of reference pose, $P_{ref}$. A similar step is used in \cite{unseen_2018_CVPR}. This provides us with a good initial image on top of which we apply the human synthesis part of our method. The body part transform process is illustrated in Fig.~\ref{fig:trans_framework} and described in detail in Sec.~\ref{body_transform}.

Fig.~\ref{fig:two-stage_framework} shows the architecture of our two-stage model. The human synthesis stage (a) takes the transformed body parts as input and produces a foreground body image, filling the ``holes'' between transformed segments and adjusting details (e.g., angle of face) to produce a photo-realistic image of the target person in the reference pose (Sec.~\ref{human_synthesis}). In the second stage, a fusion network (b) takes the generated foreground and a fixed background image as inputs and generates the final synthesized output $I_{out}$ (Sec.~\ref{fusing_network}). 
The fusion network combines the foreground and background, fixing discontinuous foreground boundaries and adding important details such as shadows. We design special training techniques and loss functions as discussed in Sec.~\ref{loss_functions}.

\vspace{-.1cm}
\subsection{Human body part transformation}
\label{body_transform}
\begin{figure}[t]
\begin{center}
\includegraphics[width=1.0\linewidth]{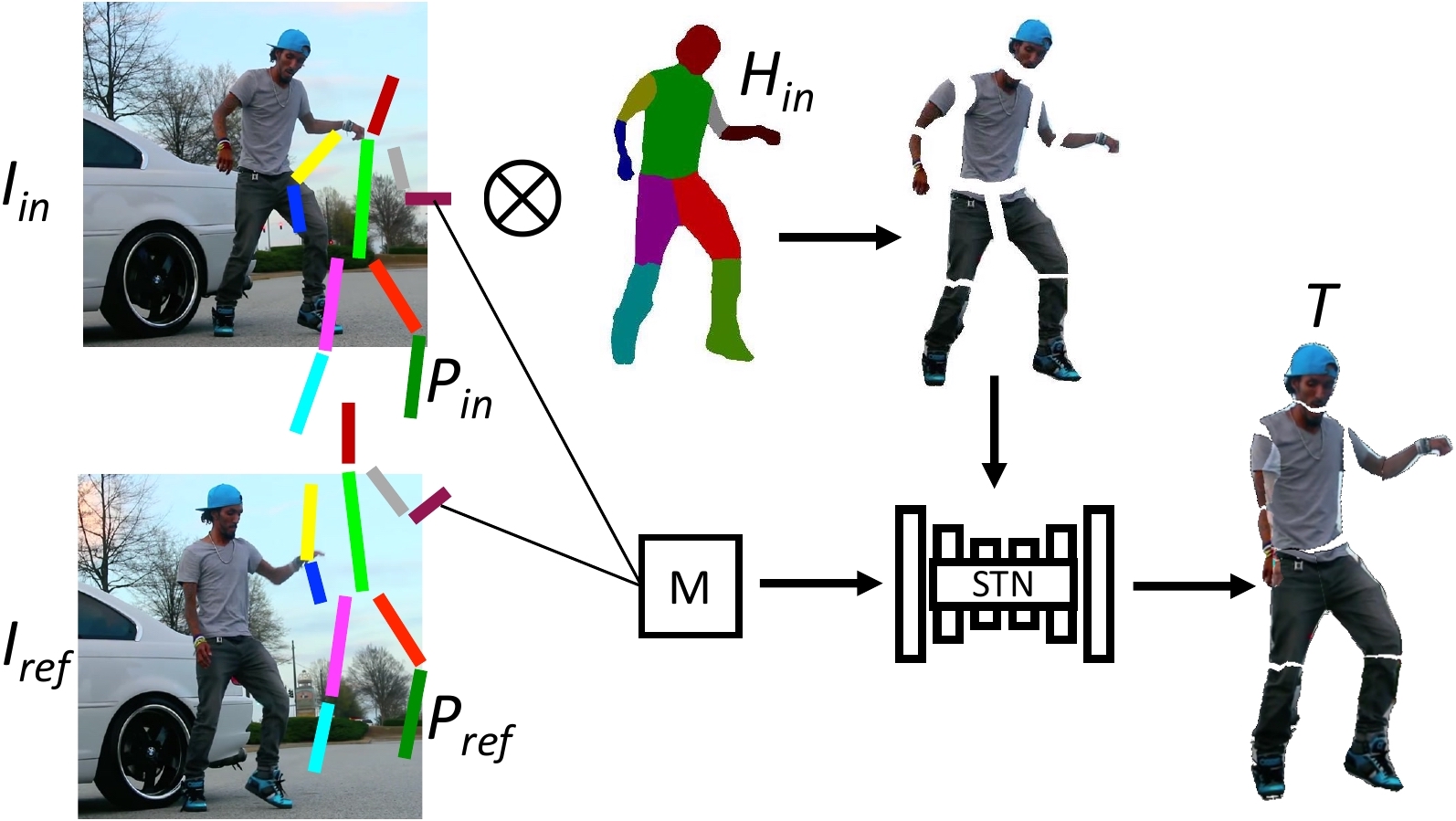}
\end{center}
\vspace{-.2cm}
   \caption{Body parts transformation. Spatial transformer network (STN) aligns the body parts of $I_{in}$ with target pose $P_{ref}$, producing body segments represented as a 3D volume $T$.} 
\label{fig:trans_framework}
\vspace{-.4cm}
\end{figure}

\vspace{-.1cm}
For each frame pair $(I_{in}, I_{ref})$, we first apply a human body parsing algorithm~\cite{WSHP} on $I_{in}$ as illustrated in Fig.~\ref{fig:trans_framework}, where $H_{in}$ is the human parsing map and $\otimes$ represents element-wise multiplication. The person from the input frame is segmented into 10 body parts (head, torso, left/right upper arms, left/right lower arms, left/right upper legs, left/right lower legs). We also employ the AlphaPose~\cite{alphapose} pose estimation algorithm to detect 2D poses, $(P_{in}, P_{ref})$, in both input and reference frames. 

Given the pose estimates, we connect line segments between pairs of key points for each of the ten body parts. Based on the corresponding line segments in $P_{in}$ and $P_{ref}$, we compute affine transformation matrices $\{M_i \in \mathbb{R}^{2 \times 3}\}_{i=1,...,10} $ which define the transformation to align each body part in $I_{in}$ with the pose in reference frame $P_{ref}$. This transformation helps normalize the body part rotation and scaling between two images (the reference person may appear smaller or larger than the target person). A spatial transformer network~\cite{STN}, which applies image warping operations including translation, scale, and rotation, is used to transform input frame body parts according to transformation matrices $M_i$ and generate reference-aligned body segments $T\in \mathbb{R}^{H{\times}W{\times}(3\times10)}$ (Fig.~\ref{fig:trans_framework}).

\vspace{-.1cm}
\subsection{Pose map representation}
\label{pose_map}
\vspace{-.1cm}
Previous works~\cite{pose_guide_2017,deformable_2018_CVPR} represent pose information as keypoint maps. We find that this can cause 'broken limbs' (missing or disconnected body parts) when synthesizing foreground human body. In our work, we alleviate this issue by encoding the position, orientation and size of each body part as Gaussian smoothed heat map parameterized by a solid circle or rectangle, as illustrated by $\textbf{P}_{ref}$ in Fig.~\ref{fig:two-stage_framework}. The shape parameters (radius, height, width, etc.) are estimated from the associated key points.


{\bf Enhancing local details:}
Besides filling holes between transformed body parts, the human synthesis model is also expected to correct the orientation of head/feet/hands since errors on these important body parts can severely degrade the perceptual quality of results. Thus, in addition to the original 10 body parts, we also add facial, hand and foot landmarks detected by OpenPose~\cite{openpose,openpose_hand}\footnote{We use AlphaPose~\cite{alphapose} as our main detection algorithm because of its robust performance, and use OpenPose~\cite{openpose} to gather additional face, hand, and foot key points which are unavailable from AlphaPose.}, and encode them in the Gaussian smoothed heat map format.
In summary, we represent the reference pose map as a 3D volume $P_{ref} \in \mathbb{R}^{H \times W \times (10 + 5)}$, with the first 10 channels representing limbs/trunk and the last 5 channels representing facial, left/right hands, left/right feet landmarks.

{\bf Temporal smoothing:} What we have described so far focuses on image-based frame-to-frame translation. Since our input pose maps are characterized by 2D spatial information, there exists ambiguity when inferring the underlying body configuration in the 3D world. Such uncertainty leads to non-smooth video outputs or temporal flickering.
To enforce smooth change across the generated frames, we introduce temporal smoothing through the inputs of our model. We include the previous $K$ reference poses as model inputs. Specifically, we stack $P_{ref}^{t-k+1}, ..., P_{ref}^{t}$ along the pose map channel where $P_{ref}^{t}$ denotes the current reference pose detected from $I^t_{ref}$. Now the stacked pose map $\textbf{P}_{ref}$ has $(10+5)K$ channels. 

\vspace{-.1cm}
\subsection{Human synthesis network}
\label{human_synthesis}
\vspace{-.1cm}
The human synthesis network generates the foreground person in the reference pose given the transformed body parts $T$ and the pose map $\textbf{P}_{ref}$ (Sec.~\ref{pose_map}).
To train the synthesis network, we convert the body parsing map $H_{ref}$ for reference image $I_{ref}$~\cite{WSHP} to a binary mask $H_{ref}^{bi}$, and extract only the foreground human region $\widetilde{I}_{ref} = I_{ref} \otimes H_{ref}^{bi}$ as training supervision. Of course, the parsed mask will not always be perfect due to algorithmic limitations. Artifacts such as broken limbs will be refined by the fusion network (Sec.~\ref{fusing_network}). Note that to avoid distraction by background clutters, the synthesis network is trained to generate human body on a green background with Chroma key compositing as shown in Fig.~\ref{fig:two-stage_framework} (a). 
This uniform background can simplify foreground extraction so that we can directly obtain the foreground mask $H_{gen}^{bi}$ of the generated frame $\widetilde{I}_{out}$ for later processing. 
As discussed in Sec.~\ref{pose_map}, the human synthesis net adjusts important pose details, such as head/feet orientations, as exemplified in Fig.~\ref{fig:example_figs} (a).

\vspace{-.1cm}
\subsection{Fusion network}
\label{fusing_network}
\vspace{-.1cm}
The fusion network is used to combine the generated human foreground image $\widetilde{I}_{out}$ with a fixed background image $B$ to generate the final output $I_{out}$. The background image is obtained by averaging all the frames in the personal video after masking out the foreground with human segmentation results~\cite{WSHP}. $B$ might contain artifacts such as blurriness due to moving objects or small camera motions that will be refined by the fusion network.

As shown in Fig.~\ref{fig:two-stage_framework} (b), the fusion network takes the combined foreground/background frame $I_{comb}$ and the stacked pose map $\textbf{P}_{ref}$ as inputs.
The combined frame $I_{comb}$ is given by: 
\vspace{-.1cm}
\begin{equation}
I_{comb} = H_{gen}^{bi} \otimes \widetilde{I}_{out} + (1 - H_{gen}^{bi}) \otimes B
\end{equation}
where $H_{gen}^{bi}$ is the foreground mask for the generated human frame $\widetilde{I}_{out}$. We apply these ``cut and paste'' operations to explicitly define the layering order of the human and background for better blending results.

\begin{figure}[t]
\begin{center}
\includegraphics[width=1.0\linewidth]{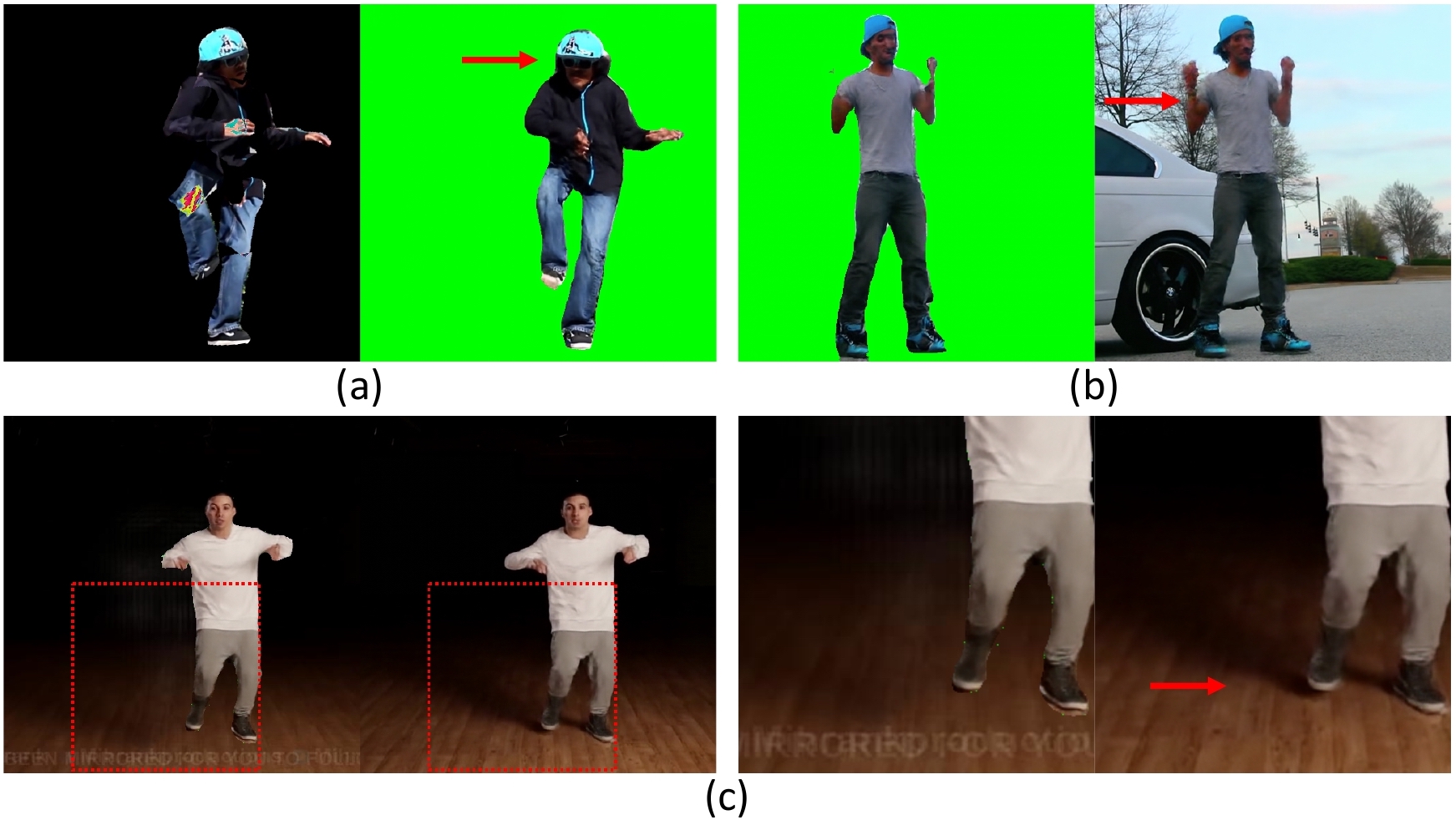}
\end{center}
\vspace{-.3cm}
   \caption{(a) The foreground synthesis net adjusts the body segments (left) to obtain correct face orientation (right). (b) The fusion net fills in the background and fixes broken limbs (e.g., arms) produced by the synthesis net. (c) The combined frame $I_{comb}$ and fusion net output $I_{out}$ (left), with their zoomed-in regions (right).}
\label{fig:example_figs}
\vspace{-.4cm}
\end{figure}

We use the reference frame $I_{ref}$ as supervision for training the fusion network. Besides blending the human with the background scene, the fusion network helps refine the human (foreground) and fills in broken limbs introduced in the previous stage as demonstrates by the example in Fig.~\ref{fig:example_figs}(b) . 
The example in Fig.~\ref{fig:example_figs}(c) shows the network's ability to add appropriate shadows for interactions between human and scene. In particular, on the left side, we show two frames $I_{comb}$ and $I_{out}$. In the zoomed-in regions on the right, we observe that the fusion network has removed artifacts from the estimated background $B$ and added shadows consistent with the lighting and person's position.

\vspace{-.1cm}
\subsection{Loss functions}
\label{loss_functions}
\vspace{-.1cm}
In order to synthesize frames at high resolution, we utilize a pix2pixHD-style~\cite{pix2pixHD} framework to train the model in a  multi-scale manner. The human synthesis network and fusion network share the same architecture as well as the same loss functions. As usual, $G$ denotes the generator and $D$ denotes the discriminator. We use $I$ to represent input frames $I_{in}$ or $I_{comb}$, $x$ to represent predicted results $\widetilde{I}_{out}$ or $I_{out}$, $y$ to represent supervision (ground truth) $\widetilde{I}_{ref}$ or $I_{ref}$, and $p$ to represent the reference pose maps $\textbf{P}_{ref}$. So we have $x = G(I, p)$. We employ several losses in our model: 

\vspace{+.15cm}
\noindent{\bf Relativistic average LSGAN loss $\mathcal{L}_{rela}$:} The traditional LSGAN~\cite{LSGAN} losses based on our task can be formatted as:
\vspace{-.1cm}
\begin{equation}
\begin{aligned}
\min_{D}\mathcal{L}_{LSGAN}(D) = & \frac{1}{2}\mathbb{E}_{y,p}[(D(y, p) - 1)^2] + \\
                                                                  & \frac{1}{2}\mathbb{E}_{x,p}[(D(x, p))^2]  \\
\min_{G}\mathcal{L}_{LSGAN}(G) = & \frac{1}{2}\mathbb{E}_{x,p}[(D(x, p) - 1)^2] 
\end{aligned}
\end{equation}
To make training more stable, we make the LSGAN framework relativistic~\cite{rela_gan}. Here, the main idea is that $D$ estimates the probability that the input is real while $G$ increases the probability that fake data is real. \cite{rela_gan} argues that it is necessary for $G$ to also decrease the probability that real data is real. 
We also apply a gradient penalty~\cite{WGAN_GP} term in $D$.
Define $\mu(.)$ as the mean operation, and the relativistic average LSGAN becomes:
\vspace{-.1cm}
\begin{equation}
\begin{aligned}
\min_{D}\mathcal{L}_{rela}(D) = & \frac{1}{2}\mathbb{E}_{x,y,p}[(D(y,p) - \mu(D(x,p)) - 1)^2] + \\
                                                                   & \frac{1}{2}\mathbb{E}_{x,y,p}[(D(x, p) - \mu(D(y,p)))^2]  + \\
                                                                   & w_{GP}\mathbb{E}_{\hat{x,p}}[(\parallel \bigtriangledown_{\hat{x,p}}D(\hat{x,p}) \parallel_{2} - 1 )^2] \\
\min_{G}\mathcal{L}_{rela}(G) = & \frac{1}{2}\mathbb{E}_{x,y,p}[(D(x, p) - \mu(D(y,p)) - 1)^2]  + \\
                                                            & \frac{1}{2}\mathbb{E}_{x,y,p}[(D(y, p) - \mu(D(x, p))^2]
\end{aligned}
\end{equation}
where $w_{GP}$ is the weight for the gradient penalty term.

\vspace{+.15cm}
\noindent {\bf Feature matching loss $\mathcal{L}_{FM}$:} We use the feature matching loss from \cite{pix2pixHD}. Specifically, we minimize the distance of the features extracted from different layers of D between real and synthesized frames:
\vspace{-.1cm}
\begin{equation}
\mathcal{L}_{FM} = \mathbb{E}_{x,y,p}\sum_{i=1}^{M}\frac{1}{N_i}[\parallel D^{i}(y,p) - D^{i}(x, p) \parallel_{1}]
\end{equation}
where $M$ is the number of layers in $D$, $N_i$ is the number of elements in each layer, and $D^i$ denotes the $i^{th}$ layer of $D$.

\vspace{+.15cm}
\noindent {\bf Perceptual loss $\mathcal{L}_{VGG}$:} We utilize the intermediate representations of VGG19~\cite{VGG} pre-trained on ImageNet~\cite{imagenet} classification. We use $\varphi(.)$ to represent the VGG network and the perceptual loss is computed as:
\begin{equation}
\mathcal{L}_{VGG} = \mathbb{E}_{x,y}\sum_{i=1}^M\frac{1}{N_i}[\parallel \varphi^{i}(y) - \varphi^{i}(x) \parallel_{1}]
\end{equation}
where $\varphi^i$ represents $i^{th}$ layer of the VGG network with $N_i$ elements and $M$ is the number of layers.

\vspace{+.15cm}
\noindent {\bf Semantic layout and pose feature losses $\mathcal{L}_{SP}$:} We apply this loss in order to encourage the model to synthesize frames with similar pose and human semantic layouts as ground truth. 
We use a pre-trained model from~\cite{mula} which jointly performs human semantic parsing and pose estimation. We extract intermediate representations from the pose encoder and parsing encoder, resulting in  feature vectors in $\mathbb{R}^{256 \times 256 \times 64}$. We use $\phi_p$ to denote the pose encoding network and $\phi_s$ to represent the semantic parsing encoding network. Then, the loss can be formatted as:
\vspace{-.1cm}
\begin{equation}
\begin{aligned}
\mathcal{L}_{SP} = & \mathbb{E}_{x,y}[\parallel \phi_{p}(y) - \phi_{p}(x) \parallel_{1}] + \\
& w_S\mathbb{E}_{x,y}[\parallel \phi_{s}(y) - \phi_{s}(x) \parallel_{1}] ,
\label{eq_sp}
\end{aligned}
\end{equation}
\vspace{-.1cm}
where $w_S$ is the weight for the semantic parsing loss.

\vspace{+.15cm}
Finally, We linearly combine all losses as: 
\vspace{-.1cm}
\begin{equation}
\begin{aligned}
\mathcal{L} =  & w_{rela}\mathcal{L}_{rela} + w_{FM}\mathcal{L}_{FM} + \\ & w_{VGG}\mathcal{L}_{VGG} + w_{SP}\mathcal{L}_{SP}
\label{eq_linear}
\end{aligned}
\end{equation}
\vspace{-.1cm}
where $w$'s represent the weights for each loss term. 

\begin{figure*}[t]
\begin{center}
\includegraphics[width=1.0\linewidth]{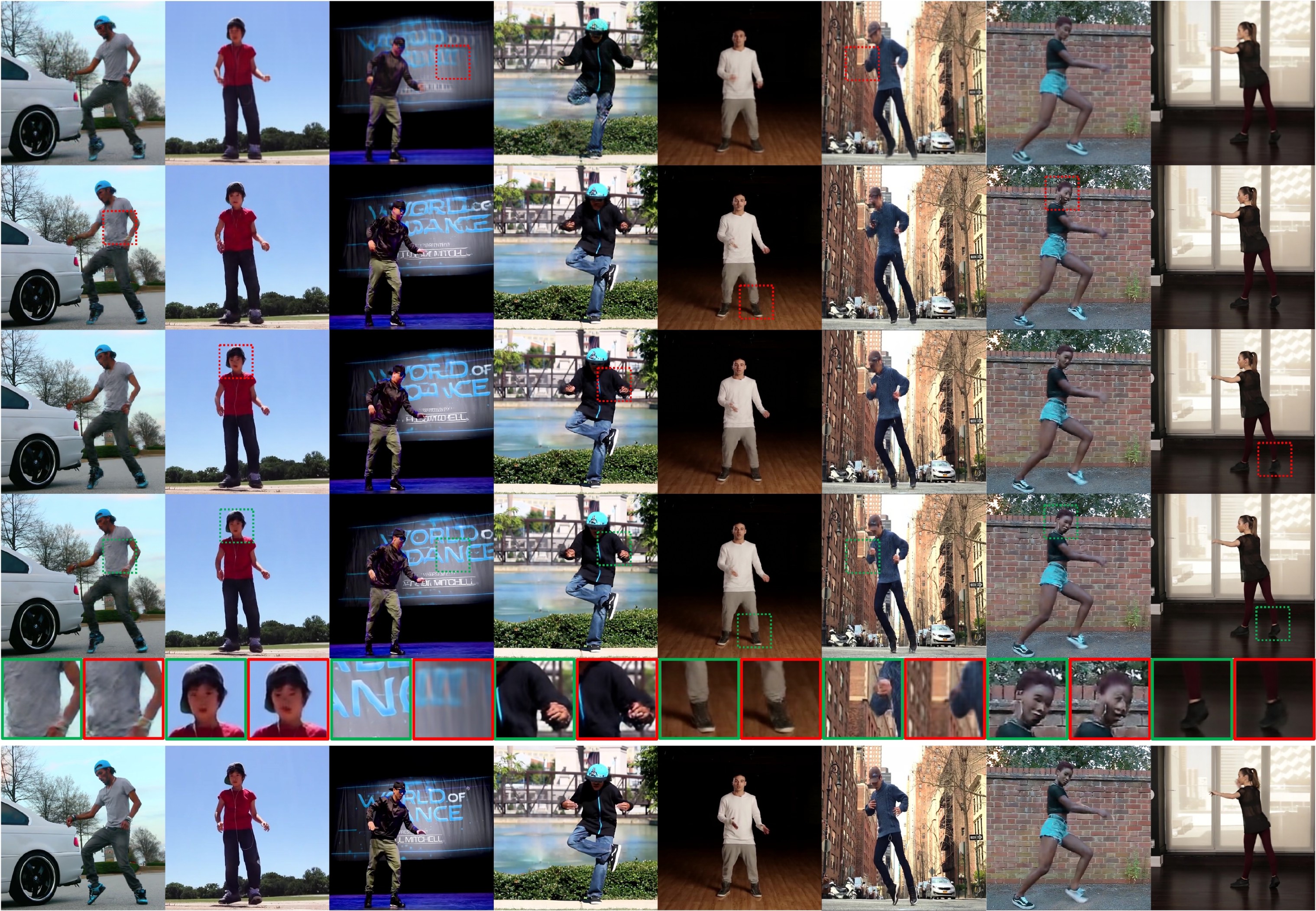}
\end{center}
\vspace{-.2cm}
   \caption{Generation results on testing set of four methods from the first to fourth row: Posewarp, pix2pixHD, Ours-baseline, Ours. Row 5 shows the zoomed-in regions and the last row shows the reference frames, which are also the ground truth outputs.}
\label{fig:test_results}
\vspace{-.3cm}
\end{figure*}

\begin{figure}[t]
\begin{center}
\includegraphics[width=1.0\linewidth]{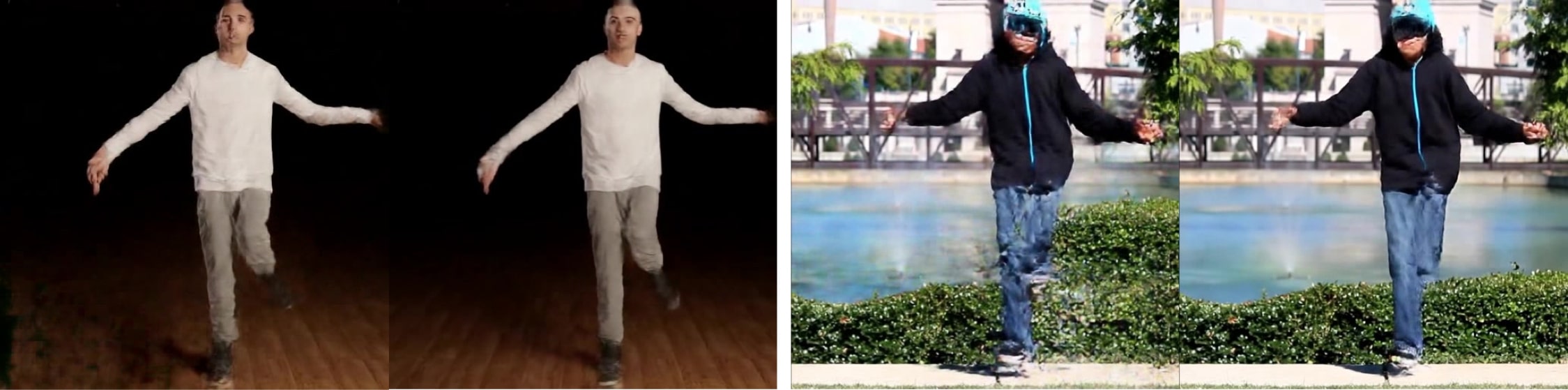}
\end{center}
\vspace{-.2cm}
   \caption{Comparison between results generated by single-stage pix2pixHD (left) and our two-stage model (right).}
\label{fig:compare}
\vspace{-.4cm}
\end{figure}

\vspace{-.1cm}
\section{Experiments}
\label{experiments}
\vspace{-.1cm}
In this section, we introduce training and inference data (Sec.\ref{personal_videos}) as well as the model architectures and training details (Sec.~\ref{model_train_details}), show qualitative visualizations (Sec.~\ref{qualitative_results}), and present numerical and human evaluations (Sec.~\ref{numerical_eval} and Sec.~\ref{human_eval}).
Additional results are presented in the supplementary material, including generated videos. 

\vspace{-.1cm}
\subsection{Dataset}
\label{personal_videos}
\vspace{-.1cm}
To study personalized motion transfer on Internet videos, we collect 8 videos from YouTube ranging from 4 to 12 minutes. Fig.~\ref{fig:example_frames} shows some example frames for each video. These personal videos include dancing videos and dancing tutorial videos. For each video, we train a personalized model.

We also collect pose reference videos from different people which are used to drive our personalized motion generation. 
16 short videos are downloaded from YouTube including various dance types such as break-dance, shuffle dance and ballet, as well as Taichi martial art which has very different motion style from the target personal videos. All video URLs are provided in the supplementary material.


\vspace{-.1cm}
\subsection{Model and training details}
\label{model_train_details}
\vspace{-.1cm}
The human synthesis network and fusion network use the same architecture. For each we apply a multi-scale generator and discriminator. Specifically, we train the model under two scales: 256$\times$256 and 512$\times$512. We use $g$ and $G = (g, \widetilde{g})$ to refer to their generators respectively. $\widetilde{g}$ means the additional convolution layers and residual blocks stacked with the start and end of $g$, and jointly trained to form $G$. We also use a two-scale discriminator for lower resolution prediction and three-scale for higher resolution. Due to GPU memory limitation, we make the number of activation maps of each layer half of the original model~\cite{pix2pixHD}. During training, we apply
Adam Stochastic Optimization~\cite{adam} with learning rate 0.0002 and batch size 4.

We set $K{=}3$ for reference pose maps $\textbf{P}_{ref}$, i.e., stacking the previous two and the current reference poses as network inputs to encourage temporal smoothness. We set the weight of the gradient penalty term $w_{GP}{=}10$ and $w_S{=}0.01$ in Eq.~\ref{eq_sp}. The linear combination weights in Eq.~\ref{eq_linear} are $w_{rela}{=}1$, $w_{FM}{=}10$, $w_{VGG}{=}10$ and $w_{SP}{=}10$.

We train one model for each personal video, sampled at 30 FPS. We split each video into training and testing sequences by ratio of 0.85:0.15 (there is no overlap between the training and testing sets) and randomly sample 20,000/2,000 $(I_{in}, I_{ref})$ frame pairs from the training/testing sub-sequence. This testing set is used for quantitative evaluation because of the availability of ground truth $I_{out}$ (which is from the same person as $I_{ref}$).

\begin{figure*}[t]
\begin{center}
\includegraphics[width=1.0\linewidth]{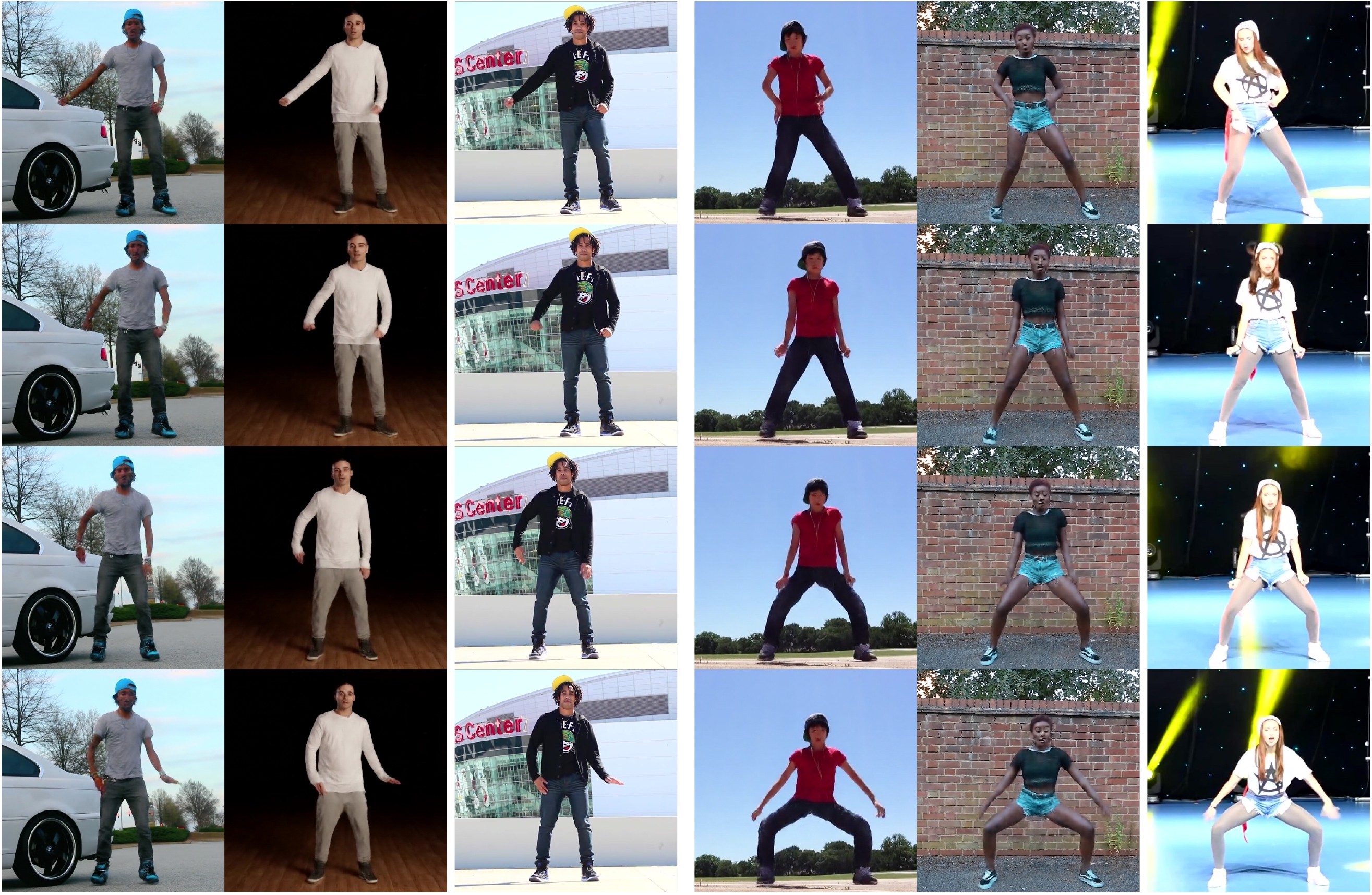}
\end{center}
\vspace{-.2cm}
   \caption{Inference generation results: we show two sets (column 1-3 and column 4-6 ) of generation results from our model. For each set, the first two columns are the synthesized results and the last column is a sequence of reference frames from a different video.}
\label{fig:infer_results}
\vspace{-.2cm}
\end{figure*}

\vspace{-.1cm}
\subsection{Qualitative results}
\label{qualitative_results}
\vspace{-.1cm}
We visualize the generated results from our model for testing and inference. For each personal video we randomly sample testing pairs that have never been seen by the model during training. In addition to using unseen frames from an input video to drive motion transfer, we can also drive motions of the target person using novel reference videos. Here there is no ground truth to compare with, but we can evaluate results using human experiments. For each reference frame from a new video, we select $I_{in}$ from the personal video with the shortest normalized pose keypoint distance to the reference pose.

We show synthesized results from our method as well as other approaches. We utilize $\textbf{pix2pixHD}$~\cite{pix2pixHD} as one baseline. This model learns a mapping directly from pose $P_{ref}$ to frame $I_{out}$. We use the loss functions and model structure from~\cite{pix2pixHD}, but reduce the number of activation maps in each layer by half (as in our model). We  also evaluate the $\textbf{Posewarp}$ model from \cite{unseen_2018_CVPR} as another baseline. Finally, we evaluate a simplified variant of our model $\textbf{Ours-baseline}$ that has the same two-stage model architecture but trained without supervision on the human synthesis result $\widetilde{I}_{out}$.
For the baseline $\textbf{Posewarp}$ method, we generate $256{\times}256$ resolution frames as in~\cite{unseen_2018_CVPR}, while results from the others are at $512{\times}512$ resolution. We show examples of both testing ($I_{ref}$ and $I_{out}$ of the same person) and inference ($I_{ref}$ and $I_{out}$ of different persons) results in Fig.~\ref{fig:test_results} and Fig.~\ref{fig:infer_results} respectively. We compare the proposed method with competing methods on the testing set and show motion transfer from novel reference videos to any of our training videos on the inference set.

\begin{figure}[t]
\begin{center}
\includegraphics[width=1.0\linewidth]{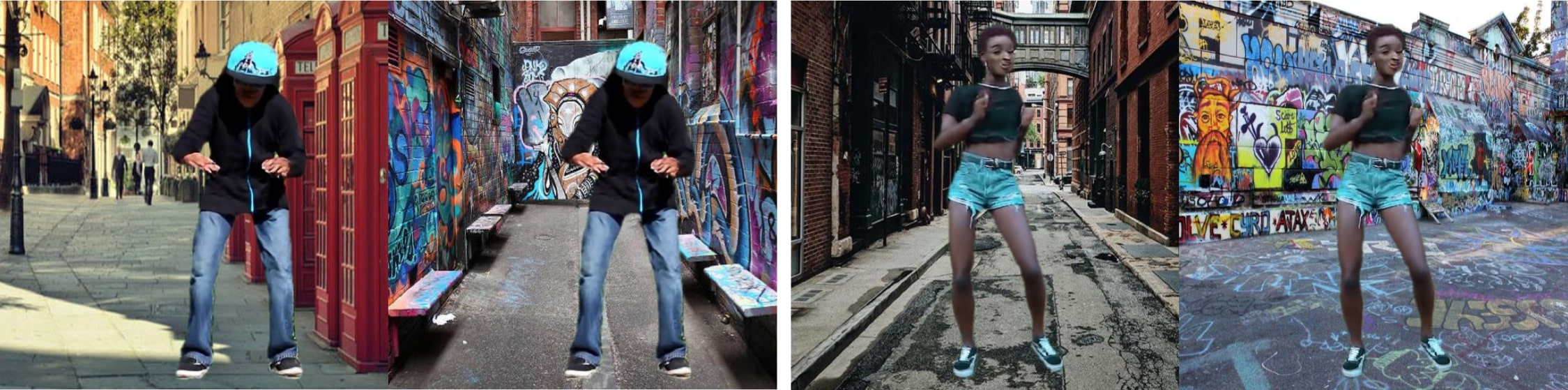}
\end{center}
\vspace{-.2cm}
   \caption{Combine generated foreground person with different scenes.}
\label{fig:random_background}
\vspace{-.4cm}
\end{figure}

Compared with single-stage models which learn direct mapping between poses and video frames, our two-stage model divides this process into easier sub-tasks. This improves generation quality for both foreground and background, and reduces blending artifacts between synthesized human and background scene as Fig.~\ref{fig:compare} shows. In this work, the human synthesis and fusion networks are jointly trained while retaining their own respective supervision. One advantage is that the intermediate human generation can be utilized to place the generated individual on new backgrounds. In fig.~\ref{fig:random_background}, we show several examples where the generated foregrounds have been composited on other scenes from the Internet. We apply simple Gaussian blur on the boundary of the foreground to alleviate border effects (more advanced blending techniques could be used to further improve visual quality).

\vspace{-.1cm}
\subsection{Numerical evaluation}
\label{numerical_eval}
\vspace{-.1cm}
We quantitatively evaluate testing results of the proposed method and other approaches with metrics Mean Square Error (MSE), PSNR, and SSIM averaged over all personal videos. 
The results for the whole frames and foreground (human) regions are reported in Table~\ref{table: scores_whole} and \ref{table: scores_foreground}, respectively
Our method achieves the best synthesis quality in MSE, PSNR, and SSIM for both full-frame and foreground-only evaluations.
The Posewarp method achieves relatively high errors in whole frame evaluation because the moving background in some videos violates the static background assumption of the method (background synthesized through hole-filling on the input frame). 

To measure the temporal coherence of generated frames, we calculate the difference between all consecutive generated frame pairs ($I_{out}^t{-}I_{out}^{t-1}$) from the testing results, and measure the MSE against the ground truth difference frame. This simple metric can be regarded as a rough surrogate to optical flow difference. As can be seen from the results averaged over all the input videos in Table~\ref{table: TC_MSE}, both our full model and ours-baseline achieve competitive temporal coherence on the synthesized results.

\begin{figure}[t]
\begin{center}
\includegraphics[width=1.0\linewidth]{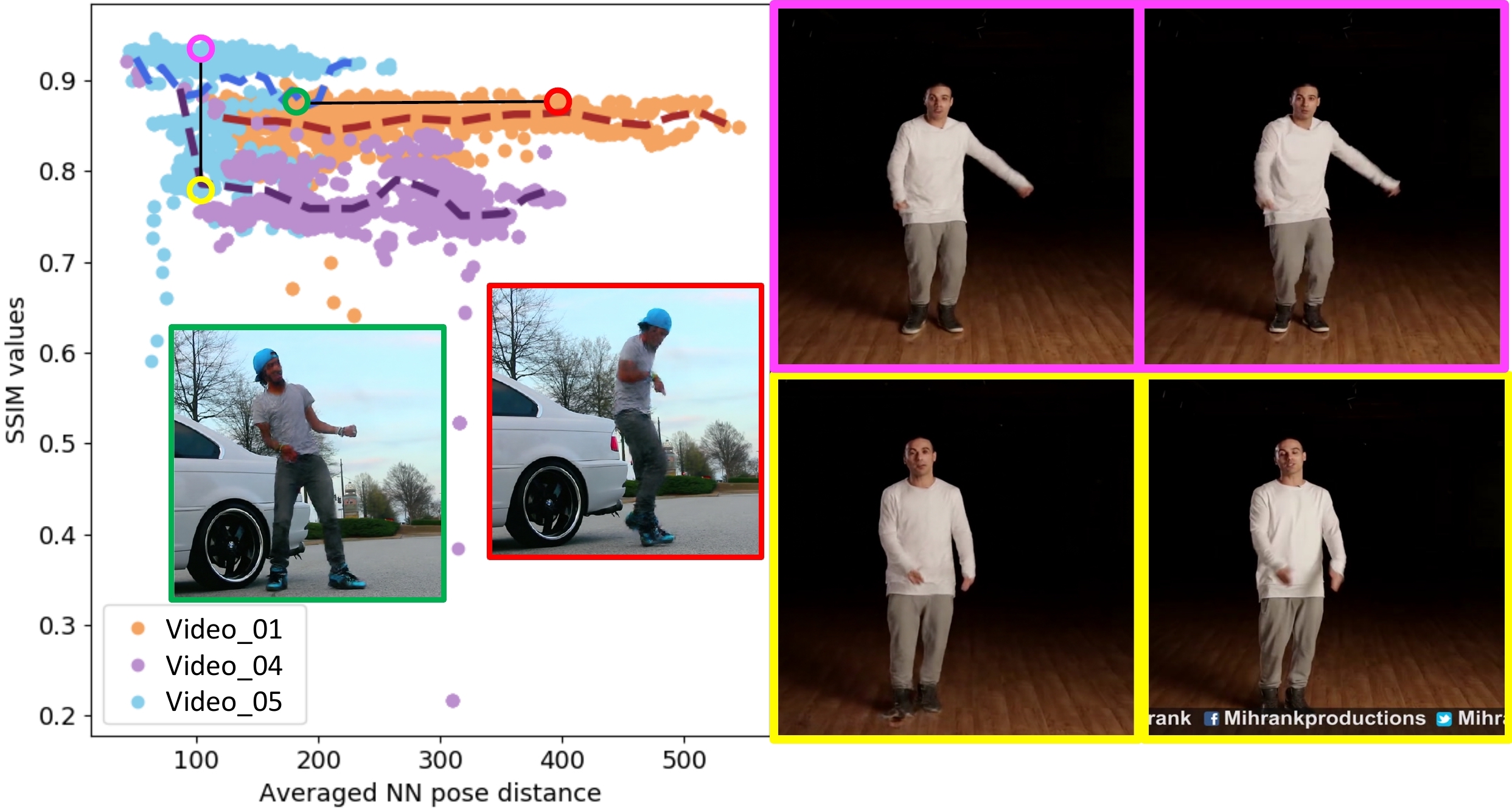}
\end{center}
\vspace{-.2cm}
   \caption{Relationship between pose distance to training samples and generation quality. The highlighted frame pair with magenta and yellow color (left is generated and right is ground truth) have the same pose distance with training sample but different SSIM scores due to the subtitle occasionally observed in the background of the yellow frame. The frame pair of green and red color have similar generation quality although in training the model has only seen similar poses as the green one but not the red one.}
\label{fig:pose_dist}
\end{figure}

Generalization to unseen reference poses, e.g. to transfer ballet motion to a target person whom has only been observed to perform hip-hop dancing, is critical for our personal model as it is trained with an Internet video with only a few minutes covering a limited portion of human pose space.
Fig.~\ref{fig:pose_dist} shows the relationship between the ``novelty'' of reference poses and the SSIM generation quality. Specifically, we compute the normalized keypoints distance between the reference pose and all training poses, and take the average of the top 10 nearest distances as the value in x axis.
The testing results from 3 personal videos are plotted as colored dots in Fig.~\ref{fig:pose_dist}. We observe that even for poses very different from training (large x axis), our model's performance remains stable. Most outliers in Fig.~\ref{fig:pose_dist} come from unexpected ground truth such as frames in the video prologue with very dark color.

\begin{table}[]
\centering
\scalebox{1.0}{
\begin{tabular}{cccc}
\hline
                    & MSE & PSNR  & SSIM \\
\hline
\hline
pix2pixHD~\cite{pix2pixHD} & 702.5714 & 20.9388 & 0.7947   \\
Posewarp~\cite{unseen_2018_CVPR} & 744.3939 & 20.6992 & 0.7663 \\
Ours-baseline & 664.7313 & 21.2318 & 0.8064  \\
Ours & 642.9080 & 21.3286 & 0.8115  \\
\hline
\end{tabular}}
\caption{Evaluation of whole frame synthesis results on testing set.}
\label{table: scores_whole}
\vspace{-.2cm}
\end{table}

\begin{table}[]
\centering
\scalebox{1.0}{
\begin{tabular}{cccc}
\hline
                    & MSE & PSNR  & SSIM \\
\hline
\hline
pix2pixHD~\cite{pix2pixHD} & 191.5246 & 25.7823 & 0.9314   \\
Posewarp~\cite{unseen_2018_CVPR} & 191.1796 & 25.8334 & 0.9264 \\
Ours-baseline & 176.9530 & 26.0477 & 0.9338  \\
Ours & 171.3259 & 26.1752 & 0.9352  \\
\hline
\end{tabular}}
\caption{Evaluation of foreground synthesis results on testing set.}
\label{table: scores_foreground}
\vspace{-.4cm}
\end{table}

\begin{table}[]
\centering
\begin{tabular}{ccc}
\hline
                    & Whole frame & Foreground   \\
\hline
\hline
pix2pixHD~\cite{pix2pixHD} & 225.7470 & 103.5557 \\
Posewarp~\cite{unseen_2018_CVPR} & 220.8329 & 103.4395 \\
Ours-baseline & 219.5680 & 98.9921   \\
Ours & 217.8404 & 99.5027   \\
\hline
\end{tabular}
\caption{The MSE of difference frame on testing set.}
\label{table: TC_MSE}
\end{table}

\vspace{-.1cm}
\subsection{Human evaluation}
\label{human_eval}
\vspace{-.1cm}
Finally, we also measure the human perceptual quality of generated results, especially for motion transferred from novel videos to a target person, where there exists no ground truth for comparison. Therefore, we conduct a human subjective test to measure motion transfer quality.

In these experiments, we compare videos generated with our proposed method, our simplified baseline model, Posewarp, and pix2pixHD by conducting a forced-choice task on Amazon Mechanical Turk (AMT). We show Turkers videos generated by each of the four methods with random order. Turkers are provided with five questions and asked to select one video as the answer to each question. The questions are related to: 1) the most complete human body (least missing body parts or broken limbs); 2) the clearest face; 3) the most isolated human and scene (the foreground and background are not mixed together); 4) the most temporally stable video (least jitters); 5) the most overall visual appealingness. For each question, we show instructional examples to help the Turker better understand the task. Three different Turkers are asked to label each group of videos for each question, and their selections are aggregated across tasks as the final result. We show the selection rate of the four methods averaged over the 8 personalized models in Table~\ref{table: human_eval}. We can see that consistent with our previous quantitative ground truth based measurements, the human evaluation also favors our proposed method on all the five questions. In particular, our proposed method achieves much better performance on body completeness, face clarity, foreground/background separation, and the overall visual appearance. For temporal stability, both our full model and baseline model outperform the existing methods.

\begin{table}[]
\centering
\begin{tabular}{ccccc}
\hline
                    & pix2pixHD & Posewarp & Ours-baseline & Ours \\
\hline
\hline
Q1 & 14.70\% & 17.59\% & 25.98\% & 41.73\% \\
Q2 & 13.39\% & 21.00\% & 25.20\% & 40.42\% \\
Q3 & 17.85\% & 22.83\% & 23.36\% & 35.96\% \\
Q4 & 19.42\% & 22.05\% & 28.87\% & 29.66\% \\
Q5 & 14.17\% & 19.69\% & 25.98\% & 40.16\% \\
\hline
\end{tabular}
\caption{Human rating percentages over different methods on complete body (Q1); clearest face (Q2); isolated foreground and background (Q3); temporal stability (Q4); and overall quality (Q5).}
\label{table: human_eval}
\vspace{-.4cm}
\end{table}

\vspace{-.1cm}
\section{Conclusion}
\label{conclusion}
\vspace{-.1cm}
In this paper, we introduced a model for personalized motion transfer on Internet videos. Our model consists of two parts: a human synthesis network and a fusion network. The former synthesizes a human foreground based on pose-transformed body parts. The latter fuses the person with a background scene and further refines the synthesis details. The evaluation shows that our method can generate personal videos of new motion with visually appealing quality and fewer artifacts than existing methods. Future directions include handling online videos with drastic camera motion (with background significantly changing from frame to frame) and motion transfer using incomplete or partially observed body parts.


{\small
\bibliographystyle{ieee}
\bibliography{egbib}
}

\end{document}